\pdfoutput=1

\documentclass[11pt]{article}
\usepackage[table,xcdraw]{xcolor}

\usepackage[final]{coling}

\usepackage{times}
\usepackage{latexsym}
\usepackage{adjustbox}

\usepackage[T1]{fontenc}

\usepackage[utf8]{inputenc}

\usepackage{microtype}

\usepackage{inconsolata}

\usepackage{booktabs}
\usepackage{multirow}
\usepackage{subfig}
\usepackage{tabularray}
\usepackage[normalem]{ulem}
\useunder{\uline}{\ul}{}
\usepackage{booktabs}
\usepackage{multirow}
\usepackage{graphicx}
\title{\textbf{\textit{\texttt{KnowledgePrompts: }}}Exploring the Abilities of Large Language Models \\ to Solve Proportional Analogies via Knowledge-Enhanced Prompting}

\author{
 \textbf{Thilini Wijesiriwardene\textsuperscript{1}},
 \textbf{Ruwan Wickramarachchi\textsuperscript{1}},
 \textbf{Sreeram Vennam\textsuperscript{2}},
 \textbf{Vinija Jain\textsuperscript{4}\thanks{Work does not relate to position at Meta.}},
\\
 \textbf{Aman Chadha\textsuperscript{3}\thanks{Work does not relate to position at Amazon.}},
 \textbf{Amitava Das\textsuperscript{1}},
 \textbf{Ponnurangam Kumaraguru\textsuperscript{2}},
 \textbf{Amit Sheth\textsuperscript{1}}
\\
 \textsuperscript{1}AI Institute, University of South Carolina, USA,
 \textsuperscript{2}IIIT Hyderabad, India \\
 \textsuperscript{3}Amazon GenAI, USA, 
 \textsuperscript{4}Meta AI, USA
\\
 \small{
   \textbf{Correspondence:} \href{mailto:thilini@sc.edu}{thilini@sc.edu}
 }
}

\begin{document}
\maketitle
\begin{abstract}
Making analogies is fundamental to cognition. Proportional analogies, which consist of four terms, are often used to assess linguistic and cognitive abilities. For instance, completing analogies like “Oxygen is to Gas as <blank> is to <blank>” requires identifying the semantic relationship (e.g., “type of”) between the first pair of terms (“Oxygen” and “Gas”) and finding a second pair that shares the same relationship (e.g., “Aluminum” and “Metal”). In this work, we introduce a 15K Multiple-Choice Question Answering (MCQA) dataset for proportional analogy completion and evaluate the performance of contemporary Large Language Models (LLMs) in various knowledge-enhanced prompt settings. Specifically, we augment prompts with three types of knowledge: exemplar, structured, and targeted. Our results show that despite extensive training data, solving proportional analogies remains challenging for current LLMs, with the best model achieving an accuracy of 55\%. Notably, we find that providing targeted knowledge can better assist models in completing proportional analogies compared to providing exemplars or collections of structured knowledge. Our code and data are available at: \url{https://github.com/Thiliniiw/KnowledgePrompts/}

\end{abstract}

\section{Introduction}

The ability to form analogies enables humans to transfer knowledge from one domain to another, making it a core component of human cognition \cite{hofstadter2001analogy, holyoak2001place, minsky1988society}. Specifically, in analogy-making, the emphasis is on the relations among objects, as it is the system of relations that is compared across domains rather than the specific objects and their attributes \cite{gentner1983structure}. Researchers have identified several types of analogies within the domain of NLP, such as proportional analogies (analogies among word/term pairs) \cite{brown1989two, chen2022kar, ushio2021bert, szymanski2017temporal, drozd2016word}, sentence-analogies \cite{jiayang2023storyanalogy, afantenos2021analogies, zhu-de-melo-2020-sentence, wang2020vector} and analogies of longer text \cite{sultan2022life, sultan2024parallelparc}. Proportional analogies, which is the focus of this paper, are presented in the form \textit{A:B::C:D}, meaning \textit{A is to B as C is to D}. These analogies involve four terms, where the relationship between the first pair of terms (A and B) is similar to the relationship between the second pair of terms (C and D).

\begin{figure*}[!ht]
\center
\includegraphics[width=\textwidth]{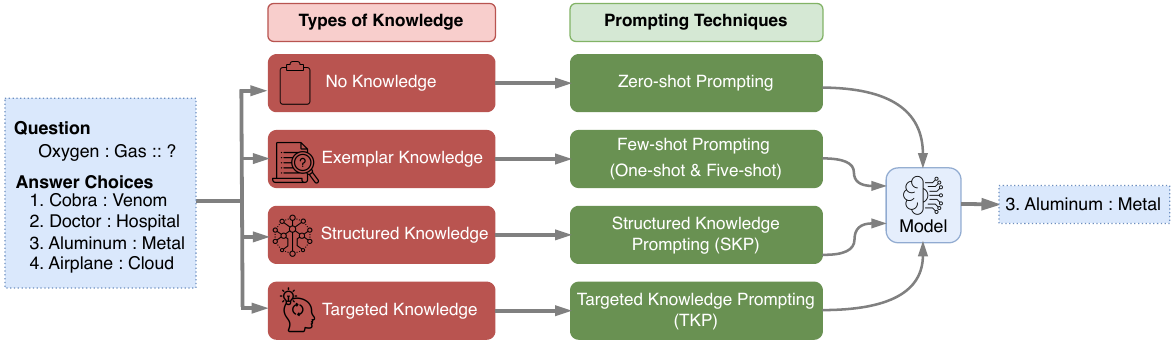}
\caption{\textbf{Knowledge-enhanced Prompting. }An illustration of our knowledge-enhanced prompting approach with types of knowledge and prompting techniques. The question consists of two terms (\texttt{``Oxygen''} and \texttt{``Gas''}), and answer choices consist of term pairs that are analogous to the question term pair. Each model is queried using the prompting techniques illustrated.}
\label{fig:approach}
\end{figure*}

Generative Artificial Intelligence (GenAI) models, particularly those recognized for their capacity to generate high-quality textual outputs\footnote{In this work, Generative AI models refer to Large Language Models (LLMs) capable of producing high-quality textual content. Therefore, we use the term ``GenAI Models'' and LLMs interchangeably.}, have emerged as a focal point of research in contemporary Natural Language Processing. The capabilities of these models are typically evaluated through a range of tasks, including question answering \cite{arora2022ask, kasai2023realtime}, reasoning \cite{zhang2024llm}, paraphrasing \cite{witteveen-andrews-2019-paraphrasing}, sentiment analysis \cite{kheiri2023sentimentgpt} and, more recently, analogical reasoning \cite{bhavya-etal-2024-anade1, wijesiriwardene-etal-2023-analogical}. Notably, \citet{wijesiriwardene-etal-2023-analogical} have demonstrated that SAT-style\footnote{SAT is a US college admission
test where proportional analogies were used to assess linguistic and cognitive abilities of examinees.} Proportional analogies pose significant challenges for LLMs, particularly when solved using intrinsic distance-based similarity measures. Conversely, \citet{webb2023emergent} have shown that GPT-3 can surpass human performance in solving proportional analogies, though these findings were based on a dataset with limited size (774 data points) and a narrow range of distinct semantic relations among term pairs (seven semantic relation types). Motivated by the need to broaden the scope of research, we scale up the evaluation by assessing a diverse set of GenAI models on a larger, more comprehensive proportional analogy dataset. Additionally, we employ various prompting techniques enhanced with multiple types of knowledge to understand model capabilities in completing proportional analogies.

Our primary contribution lies in conducting a comprehensive evaluation of nine GenAI models, specifically assessing their performance in solving proportional analogies presented in a multiple-choice format. Considering the limitations of existing proportional analogy datasets, which typically comprise fewer than a thousand data points and a restricted range of relation types, we present a substantially larger dataset. Our dataset contains 15K proportional analogies with 236 distinct relation types. We evaluate the nine GenAI models on the 15K dataset using four distinct prompting techniques: (i) \textbf{Zero-shot Prompting}, where no additional knowledge is incorporated into the prompt, (ii) \textbf{Few-shot Prompting}, where exemplar knowledge in the form of examples from the dataset is included in the prompt, (iii) \textbf{Structured Knowledge Prompting (SKP)}, where the prompt is augmented with structured knowledge in the forms of lexical, commonsense, and world knowledge drawn from WordNet \cite{mccrae-etal-2019-english}, ConceptNet \cite{speer2017conceptnet}, and Wikidata \cite{vrandevcic2014wikidata} respectively and (iv) \textbf{Targeted Knowledge Prompting (TKP)}, which integrates targeted knowledge in the form of specific semantic relationships necessary for solving proportional analogies, along with the cognitive process behind such reasoning. To the best of our knowledge, this study is the first to explore knowledge-enhanced prompting strategies for solving proportional analogies.

Our findings indicate that completing proportional analogies is highly challenging for current LLMs and incorporating targeted knowledge significantly enhances model performance, with the best-performing model showing an improvement of approximately +21\% compared to prompts without any knowledge, and around +45\% relative to prompts enhanced with structured knowledge. The underperformance of SKP relative to Zero-shot Prompting suggests that the mere inclusion of relevant knowledge may not always improve model performance.

\section{Related Work}

In this section, we introduce related literature on
the main topics of our paper: proportional analogies and LLMs, prompting techniques, and knowledge-enhancement in LLM prompting.

\subsection{Proportional Analogies and LLMs}
One of the earliest methods for solving proportional analogies was Latent Relational Analysis (LRA), introduced by \citet{turney2005measuring}. LRA determines analogy by measuring the similarity in semantic relationships shared between word pairs, considering them analogous if they exhibit a high degree of relational similarity. With the advent of neural networks, vector difference-based methods \cite{vylomova-etal-2016-take, allen2019analogies, mikolov2013linguistic} were used to address proportional analogies. As LLMs based on the Transformer architecture \cite{vaswani2017u} gained prominence, researchers began investigating the potential of LLMs, particularly Generative Artificial Intelligence (GenAI) models, for solving proportional analogies \cite{brown2020language,ushio2021bert, webb2023emergent}. Specifically, \citet{webb2023emergent} demonstrated strong performance using a single model (GPT-3) on four relatively small proportional analogy datasets. Our study extends this work by scaling up the evaluation to a substantially larger dataset and by assessing nine contemporary GenAI models across six distinct prompting approaches. Additionally, we introduce a novel exploration of the impact of incorporating various types of knowledge when evaluating GenAI models on proportional analogies.

\subsection{Prompting and Knowledge-enhanced Prompting}

GenAI models are built on LLMs that are trained on extensive datasets and optimized for various tasks, including question-answering. 
This training implies that these models encapsulate the knowledge in the data, allowing them to effectively answer natural language queries \cite{roberts-etal-2020-much, zhu2023physics}. Prompting involves transforming an input query into a structured natural language statement (prompt) and presenting it to the model, which then guides the output generation process of the model.  \cite{schulhoff2024prompt, hadi2023large, liu2023pre}. Generating outputs through prompting requires only forward passes during inference time, without any weight updates. Prompts can be created either manually \cite{wei2022chain, schulhoff2024prompt} or automatically \cite{ye2023prompt,10.1145/3411763.3451760,deng-etal-2022-rlprompt}; in this work, we employ the more intuitive manual approach.

Prompts can be categorized based on the context they provide. Zero-shot prompts \cite{brown2020language} contain only instructions related to solving a specific task, whereas Few-shot prompts \cite{brown2020language} include both the instructions and one or more examples. Providing examples when querying models is a paradigm broadly known as  In-context Learning (ICL) \cite{brown2020language}. Chain-of-Thought (CoT) Prompting is designed to guide models through the reasoning process required to solve a task by presenting an exemplar that includes the question, reasoning path, and correct answer \cite{wei2022chain} or by just incorporating a thought-inducing phrase such as ``Let's think step by step'' \cite{kojima2022large} (Zero-shot-CoT). Unlike conventional CoT prompting, which often includes an exemplar, our adaptation termed TKP does not provide an exemplar. Instead, it enhances the prompt with the targeted knowledge specific to solving proportional analogies. As a result, TKP is more akin to Zero-shot-CoT \cite{kojima2022large} than to traditional CoT \cite{wei2022chain}.

The enhancement of LLM performance through the integration of external knowledge, both unstructured and structured, has been extensively studied \cite{yu2022survey}. Some approaches transform external knowledge from multiple documents into graph structures and utilize these graphs to enhance LLM querying \cite{wang2024knowledge}. Additionally, some methods directly employ structured knowledge \cite{baek-etal-2023-knowledge}. Retrieval-augmented generation (RAG) has recently emerged as an umbrella term encompassing all these techniques, where user queries are enriched with content retrieved from external sources to enhance model performance \cite{lewis2020retrieval, ding2024survey, mialon2023augmented, schulhoff2024prompt}. In this work, we utilize multiple types of knowledge, including targeted and structured knowledge (from three sources), to assess the impact on LLM performance in solving proportional analogies. To the best of our knowledge, this is the first study to explore the capabilities of LLMs in solving proportional analogies using knowledge-enhancement approaches.

\section{Approach}

As illustrated in Figure \ref{fig:approach}, given a proportional analogy MCQ where the question consists of a single term pair \texttt{(e.g., ``Oxygen'' and ``Gas'')}, the GenAI model is required to provide the correct answer choice from five, four or three choices. \textbf{Zero-shot Prompting}, only include the MCQ and a simple instruction on how to produce the output without any knowledge enhancement added to the prompt. Next, we enhance the Zero-shot Prompt with exemplars of solved MCQs from the dataset. We consider this approach as enhancing the prompt with ``exemplar knowledge'' and refer to this prompting technique as \textbf{Few-shot Prompting}. We experiment with one exemplar (One-shot Prompting) and five exemplars (Five-shot Prompting). Then a combination of lexical, commonsense, and world knowledge from structured sources—WordNet, ConceptNet, and Wikidata, respectively—is added to the Zero-shot Prompts for knowledge enhancement, resulting in what we call \textbf{SKP}. Finally, the zero-shot prompt is enhanced with targeted knowledge and we identify this prompting technique as \textbf{TKP.} Targeted knowledge is composed of, the semantic relationship shared between the question term pair and the cognitive process behind solving the proportional analogy. We detail the prompting techniques in Section \ref{prompting_techniques}.

\subsection{Dataset Creation}
\label{sec:dataset}
We introduce a 15K dataset of proportional analogies containing 5-way, 4-way and 3-way MCQs. Table \ref{tab:data-stats} presents the dataset statistics along with examples from the dataset. We generate 14K questions out of the 15K based on the work by \cite{yuan2023analogykb}. \citet{yuan2023analogykb} introduced an automatically generated million-scale analogy knowledge base based on ConcepNet and Wikidata knowledge graphs. \citet{yuan2023analogykb} acquire analogies of the same relations directly utilizing the concept pairs in the above-mentioned knowledge graphs. To acquire analogies of analogous relations (analogies consisting of two concept pairs with two relations that are analogous to each other) \citet{yuan2023analogykb} utilize the in-context learning abilities of LLMs.
We adopt this resource (specifically the analogies of same relations) to develop n-way \texttt{(n=[3, 4, 5])} MCQs as follows. A single n-way MCQ consist of a pair of terms representing the \textit{question} and \texttt{n} term pairs representing the \textit{answer choices}, among which only one term pair is the correct answer. The semantic relationship between the term pair in the question is the same as the semantic relationship shared between the term pair which is the correct answer. The rest of the incorrect answer choices consist of term pairs with different semantic relationships among them. 

\begin{figure*}
    \centering
    \includegraphics[width=\textwidth]{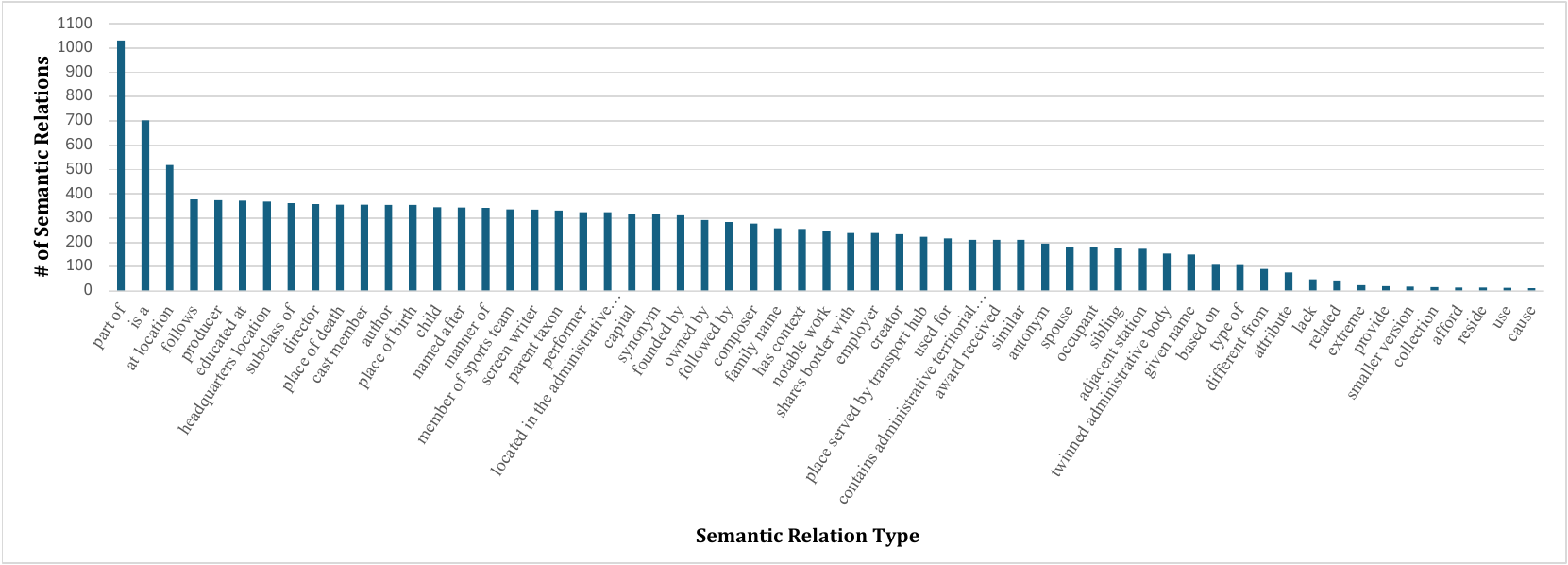}
    \caption{\textbf{Distribution of Semantic relations.} The distribution of the top 59 semantic relations (these are the frequencies of semantic relations between the question word pair )}
    \label{fig:rel_dist}
\end{figure*}

Thousand data points out of the 15K are borrowed from work by \citet{ushio2021bert,turney2003combining,boteanu2015solving}\footnote{Unlike the 14K MCQs created based on AnalogyKB, these 1K data points do not provide the semantic relationship shared between the question term pair explicitly, therefore we employ two NLP researchers to discuss and manually identify the shared semantic relationship.} and contain 5-way, 4-way and 3-way MCQs. We highlight that, compared to previous proportional analogy MCQ datasets used for research \cite{webb2023emergent, ushio2021bert}, the current dataset provides a significant increase in question quantity ($\sim$15-times) and diversity (with respect to the diversity of semantic relations between terms). Our dataset also includes the semantic relationship shared by the question term pair compared to other datasets that do not include this information \cite{ushio2021bert,turney2003combining,boteanu2015solving}. Our dataset contains 59 semantic relationship types with more than 10 instances each. The distribution of these relationships (focusing on the top 59 types) is depicted in Figure \ref{fig:rel_dist}.

\begin{table*}[]
\footnotesize
\resizebox{\linewidth}{!}{
\begin{tabular}{@{}l|l|l|l||lr@{}}
\toprule
\multicolumn{4}{c||}{\textbf{Questions}} & \multicolumn{2}{c}{\textbf{Relations}} \\ \midrule
\textbf{Question Type (MCQ)} & 5-way  & \multicolumn{1}{l|}{4-way} & 3-way  & Top 5 Relation Types & \# Data Points \\ \midrule
\multirow{4}{*}{\hspace{0pt}\textbf{Example}} & \multirow{4}{*}{\begin{tabular}[c]{@{}l@{}}\hspace{0pt}\textbf{\textit{Question}}: Tenable: Indefensible\\ \textbf{\textit{Choices}}: \\ ~~(1) Unique : Unprecedented \\ ~~(2) Dire : Pressing\\ ~~(3) Bleak : Desolate \\ ~~(4) Theoretical : Concrete\\ ~~(5) Recondite : Scholarly~\end{tabular}} & \multicolumn{1}{l|}{\multirow{4}{*}{\begin{tabular}[c]{@{}l@{}}\hspace{0pt}\textbf{\textit{Question}}: Haiku: Poem\\ \textbf{\textit{Choices}}:\\ ~ ~(1) Song : Musician\\ ~ ~(2) Novel : Book\\ ~ ~(3) Artist : Painting\\ ~ ~(4) Page : Typeface\end{tabular}}} & \multirow{4}{*}{\begin{tabular}[c]{@{}l@{}}\hspace{0pt}\textbf{\textit{Question}}:Ancient: Old\\ \textbf{\textit{Choices}}:\\ ~ ~(1) Crazy : Unhealthy\\ ~ ~(2) Delicious : Tasty\\ ~ ~(3) Smart : Intelligent\end{tabular}} & \\
 &  & \multicolumn{1}{l|}{} &  & ~ ~ part of & 1030 \\
 & &  \multicolumn{1}{l|}{} &  &  ~ ~ is a & 702 \\
 &  & \multicolumn{1}{l|}{} &  & ~ ~ at location & 518 \\
 &  & \multicolumn{1}{l|}{} &  & ~ ~ follows & 376 \\
 &  & \multicolumn{1}{l|}{} &  & ~ ~ producer & 374 \\ 
  &  & \multicolumn{1}{l|}{} &  &  \\ \midrule
\multicolumn{1}{l|}{\textbf{Amount~}} & \multicolumn{1}{c|}{14386} & \multicolumn{1}{c|}{610} & \multicolumn{1}{c||}{4} & \multicolumn{1}{l}{} \textbf{Total \# relation types} &  \textbf{236} \\ \bottomrule
\end{tabular}
}
\caption{\textbf{Dataset statistics}. The dataset consist of 15K MCQs that share 236 semantic relation types among them.}
\label{tab:data-stats}
\end{table*}

\subsection{Model Details}

GenAI models are designed to generate content that are often indistinguishable from human-produced output. Current state-of-the-art GenAI models are largely based on the Transformer architecture \cite{10.5555/3295222.3295349}. In this work we compare the following popular open-source and proprietary GenAI models for their ability to solve proportional word analogy MCQs by incorporating variety of knowledge: (i) Falcon, a causal decoder-only model \cite{almazrouei2023falcon}, (ii) FlanT5 \cite{pmlr-v202-longpre23a}, a T5 \cite{10.5555/3455716.3455856} based model trained on the Flan collection of datasets, (iii) GPT2 \cite{Radford2019LanguageMA}, the first series of models to popularize in-context instructions, (vi) Mistral \cite{jiang2023mistral}, leveraging transformers architecture \cite{10.5555/3295222.3295349} with several new introductions such as sliding window attention and pre-fill chunking, (v) Orca \cite{mukherjee2023orca}, based on LLaMA model family \cite{touvron2023llama} and fine-tuned on complex explanation traces obtained from GPT-4 \cite{achiam2023gpt}, (vi) Zephyr \cite{tunstall2023zephyr}, a fine-tuned version of Mistral trained on public datasets and optimized with knowledge distillation techniques. (vii) CodeT5 \cite{wang2021codet5}, a unified pre-trained encoder-decoder transformer model leveraging code semantics and finally (viii) CodeParrot \cite{noauthor_codeparrot_2023}, a model based on GPT-2 and trained to generate python code (ix) GPT-3.5-Turbo \footnote{\url{https://platform.openai.com/docs/models/gpt-3-5-turbo}}. Further details of the models used are presented in Appendix \ref{sec:model_details}. 

\subsection{Prompting Techniques} \label{prompting_techniques}
Currently, the most popular approach to Multiple Choice Question Answering (MCQA) is via cloze-style prompting \cite{brown2020language,robinson2023leveraginglargelanguagemodels} where each answer choice is concatenated to the question separately and  scored independently by the language model (LM). This style of prompting is problematic since it prevents the LM from comparing and contrasting all available options simultaneously. Additionally, it is computationally expensive, as it requires multiple forward passes through the LM to identify the correct answer \cite{robinson2023leveraginglargelanguagemodels}. To address these limitations, we adopt the prompt phrasing introduced by \citet{robinson2023leveraginglargelanguagemodels} with task-specific modifications. Specifically, the question and its symbol-enumerated candidate answers are provided to the model as a single prompt. \citet{robinson2023leveraginglargelanguagemodels} do not include specific instructions in the prompt for the model to output only the choice symbol. But we observe that adding such specific instructions reduce the model hallucinations. Therefore we use specific, non-ambiguous language to instruct the model to only output the relevant choice symbol. The prompting techniques are detailed below (See example prompts in appendix \ref{sec:prompts}). 

\subsubsection{Zero-shot Prompting}
In Zero-shot Prompting, the question, all multiple choice answers and  the instructions are provided in natural language (no knowledge is provided).

\subsubsection{Few-shot Prompting}
We demonstrate the task to the model by providing several exemplars in the form of question, answer choices and the correct answer choice. Then the actual question and answer choices are provided requiring the model to choose the correct answer choice. We employ one-shot and five-shot prompting under the few-shot prompting strategy where one example and five examples are provided respectively. We select these quantities of exemplars to strike a balance between the models' maximum accepted context length and the computational resources required. To obtain the exemplars, we employ a semantic similarity based filtering mechanism as follows. We encode each proportional analogy MCQ in the dataset using a SOTA sentence encoding transformer model\footnote{\url{https://huggingface.co/sentence-transformers/all-mpnet-base-v2}}, and identify the most semantically similar single example/ five examples based on Cosine similarity.

\subsubsection{Structured Knowledge Prompting (SKP)}
We retrieve knowledge from structured sources, filter it, and then integrate the resulting refined knowledge into the prompts. We detail this process in the subsequent sections.

\paragraph{Knowledge Retrieval.}
We leverage the following widely-used large knowledge sources to obtain three types of knowledge: (i) Wikidata \cite{vrandevcic2014wikidata}, which provides world knowledge in the form of explicit information about specific instances, encompassing billions of nodes and edges \cite{wang2021cognet}; (ii) ConceptNet \cite{speer2017conceptnet}, a general-domain commonsense knowledge graph with 799,273 nodes and 2,487,810 edges; and (iii) WordNet \cite{mccrae-etal-2019-english}, a lexical database for the English language containing 161,338 words, 120,135 synsets, and 415,905 semantic relations.  

We retrieve knowledge from above sources as follows. Since analogies focus on relations oppose to entities or entity attributes \cite{gentner1983structure}, when retrieving knowledge from knowledge sources we focus on path finding approaches oppose to sub-graph extraction approaches. To extract both world and commonsense knowledge, we utilize the path-finding approach by \citet{lin-etal-2019-kagnet} that identifies connections between each term pair (in both the question and answer choices). Specifically, we extract paths of length \textit{k}\footnote{\textit{k} is set to 2 for Wikidata and 3 for ConceptNet, as longer paths tend to introduce excessive noise and reduce efficiency.} from ConceptNet and Wikidata. When retrieving lexical knowledge from WordNet, we extract the shortest path between term pairs.

\begin{figure}
    \centering
    \includegraphics[width=\linewidth]{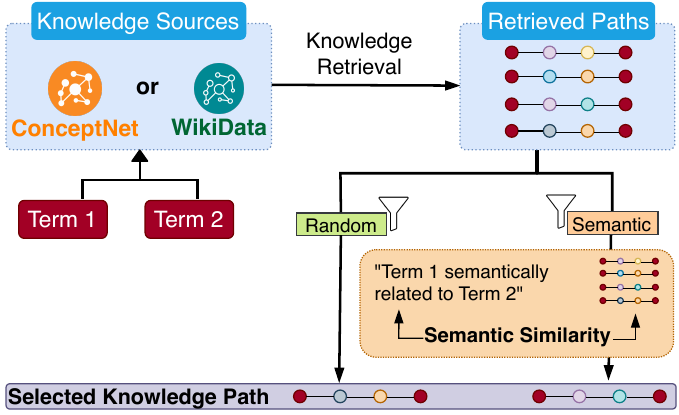}
    \caption{\textbf{An illustration of the knowledge filtering approach.} ``Random'' indicates Random Filtering and ``Semantic'' indicates Semantic Filtering.} 
    \label{fig:knowledge-filteirng}
\end{figure}

\paragraph{Knowledge Filtering.} 
For each term pair in the question and answer choices, multiple knowledge paths may be retrieved. To ensure the prompts stay within the maximum context length limit of the evaluated language models, we filter the retrieved paths and retain a single path for Wikidata and ConceptNet (See Figure \ref{fig:knowledge-filteirng}). Filtering is not performed on WordNet since a single path (shortest) is always retrieved.

The filtering mechanisms we employ are as follows: (i) \textbf{Random Filtering}, where one path is randomly selected from the list of available paths; and (ii) \textbf{Semantic Filtering}, which selects the path most semantically similar to the term pairs. The term pairs (in question and answer choices) are formatted to ``term pair sentences'' in the following form \textsc{<term\_1> is semantically related to <term\_2>} and returned paths are also formatted to ``path sentences'' in the form of \textsc{[<node1\_name> <relation1\_name> <node2\_name>, <node2\_name> <relation2\_name> <node3\_name>, ...]}. Both term pair sentences and path sentences are then encoded using a SOTA sentence encoding transformer model\footnote{\url{https://huggingface.co/sentence-transformers/all-mpnet-base-v2}} and the path sentence with the highest cosine similarity to term pair sentence is filtered as relevant knowledge and referred to as \textit{knowledge paths}\footnote{specific format of Wikidata knowledge paths is \texttt{[<node1\_name> <relation1\_name> <node2\_name>, <node2\_name> <relation2\_name> <node3\_name>]} and ConceptNet knowledge path is \texttt{[<node1\_name> <relation1\_name> <node2\_name>, <node2\_name> <relation2\_name> <node3\_name>, <node3\_name> <relation3\_name> <node4\_name>]}}. 

\paragraph{Generating Prompt.} The filtered knowledge paths are appended to the zero-shot prompt after the question and the answer choices to create the SKP and the model is instructed to use the knowledge if necessary. Based on the knowledge filtering mechanism SKP can be referred to as SKP[random] or SKP[semantic].

\subsubsection{Targeted Knowledge Prompting (TKP)}
When solving proportional analogies, humans typically examine the question term pair, identify the semantic relationship between the two terms, and select the answer pair that shares the same or a similar relationship. Inspired by this cognitive process, we modify the traditional Chain-of-Thought (CoT) prompting technique \cite{wei2022chain} to provide the model with ``targeted knowledge'' in the form of  (i) semantic relationship shared by the question term pair (ii) cognitive process used by humans when evaluating such analogies, via the prompt. 

\section{Experimental Setting}
We have conducted a comprehensive set of experiments across nine GenAI models over six prompt variants on a 15K dataset, totalling to 54 ($9 X 6$) experiments. The implementation details are included in Appendix \ref{sec:implementation_details}

\begin{table*}[!ht]
\footnotesize
\resizebox{\linewidth}{!}{%
\begin{tabular}{l|c|cc|cc|c}
\toprule
\multicolumn{1}{l|}{\multirow{2}{*}{\textbf{Model Name}}} &
  \multirow{2}{*}{\textbf{Zero-shot Prompting}} &
  \multicolumn{2}{c|}{\textbf{Few-shot Prompting}} &
  \multicolumn{2}{c|}{\textbf{Structured Knowledge Prompting \vspace{0.25em} }}  &
  \multirow{2}{*}{\textbf{Targeted Knowledge Prompting}} \\ 
\multicolumn{1}{c}{} &                & \multicolumn{1}{c}{\textbf{One-shot}}       & \textbf{Five-shot }     & \multicolumn{1}{c}{\textbf{Random}} & \textbf{Semantic }   &                \\ \midrule
Falcon                 & 24.17          & \multicolumn{1}{c}{23.21}          & 22.61          & \multicolumn{1}{c}{24.75}  & {\ul 25.01} & \textbf{25.4}  \\ 
FlanT5                 & 36.47          & \multicolumn{1}{c}{{\ul 40.09}}    & 38.07          & \multicolumn{1}{c}{14.43}  & 14.62       & \textbf{44.26} \\ 
GPT2                   & \textbf{22.65} & \multicolumn{1}{c}{{\ul 22.49}}    & 7.19           & \multicolumn{1}{c}{6.29}   & 6.17        & 21.64          \\ 
Mistral                & 26.59          & \multicolumn{1}{c}{26.22}          & {\ul 27.34}    & \multicolumn{1}{c}{24.58}  & 24.42       & \textbf{27.37} \\ 
Orca                   & \textbf{24.54} & \multicolumn{1}{c}{23.28}          & 14.11          & \multicolumn{1}{c}{18.48}  & 18.81       & {\ul 24.2}     \\ 
Zephyr                 & 29.46          & \multicolumn{1}{c}{{\ul 34.05}}    & \textbf{35.87} & \multicolumn{1}{c}{16.13}  & 17.22       & 15.83          \\ 
CodeT5                 & 20.64          & \multicolumn{1}{c}{\textbf{24.33}} & 0              & \multicolumn{1}{c}{16.15}  & 17.47       & {\ul 21.64}    \\ 
CodeParrot             & {0}  & \multicolumn{1}{c}{{\ul 10.11}}    & \textbf{12.6}              & \multicolumn{1}{c}{0}      & 0.01        & 2.09           \\ 
GPT-3.5-Turbo         & {\ul 45.7}     & \multicolumn{1}{c}{31.79}          & 41.21          & \multicolumn{1}{c}{38.29}  & 38.79       & \textbf{55.25} \\ \bottomrule
\end{tabular}%
}
\caption{\textbf{MCQ Performance of models.} Performance is reported in EMA percentage. Best performance of each model is indicated in \textbf{bold} and the second best performance is indicated by \ul{underline}.}
\label{tab:main_results}
\end{table*}

\section{Results and Discussion}
Proportional analogy multiple-choice questions (MCQs) are presented to each GenAI model using the previously described prompts. The model's response is extracted from the generated output, and accuracy is measured using Exact Match Accuracy (EMA) \cite{rajpurkar2016squad}. While more flexible evaluation metrics such as BLEU \cite{papineni2002bleu} and ROUGE \cite{lin2004rouge} are commonly used to assess GenAI-generated outputs, we employ EMA because MCQs are inherently evaluated in a binary manner, where partial correctness is not rewarded. We report EMA as a percentage for each model and prompt variant. The results are presented in Table \ref{tab:main_results}. 

\subsection{Model Performance and Prompting Techniques}

\begin{figure}
    \centering
    \includegraphics[width=\linewidth]{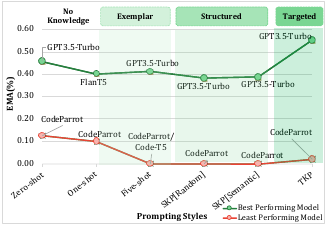}
    \caption{\textbf{Best and least performing models} for each prompting technique.}
    \label{fig:results}
\end{figure}

The highest overall performance was attained by GPT-3.5-Turbo, achieving an EMA of 55.25\%. This result underscores the challenge that proportional analogies pose for current state-of-the-art GenAI models. This accuracy was obtained through Targeted Knowledge Prompting where the prompt was enhanced with targeted knowledge (See Figure \ref{fig:results}). Interestingly, the same model, when enhanced with structured knowledge, underperformed with an accuracy of ~38\% (EMA for SKP[random] is 38.29\% and SKP[semantic] is 38.79\%), compared to Zero-shot prompting (EMA 45.7\%). This suggests that simply adding knowledge, even from diverse sources, may not be beneficial for cognitively demanding tasks such as proportional analogy completion. 
Out of the nine models four (Falcon, Flan, Mistral and GPT-3.5-Turbo) performs the best when prompted with Targeted Knowledge Prompts and two (GPT2 and Orca) performs the best with Zero-shot prompts with no knowledge enhancement. CodeT5 performs the best with one-shot prompts and Zephyr and CodeParrot performs the best with five-shot prompts.
We also observe that models trained specifically on code generation such as CodeT5 and CodeParrot (specially CodeParrot) perform at the lower end of the spectrum despite the demonstrated abilities of them to perform well on other MCQ datasets \citet{robinson2023leveraginglargelanguagemodels}. We believe this is due to the challenging nature of the proportional analogy completion task.

\begin{figure*}
    \centering
    \includegraphics[width=\textwidth]{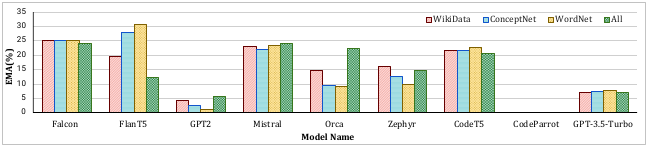}
    \caption{\textbf{Perfromance with structured knowledge. }Performance of each model when Structured Knowledge Prompting with semantic filtering (SKP[semantic]) is used. \textbf{All} indicates the prompt is enhanced with all three types of knowledge (Wikidata, ConceptNet and WordNet). EMA values are reported on ~20\% of the 15K dataset where all three knowledge types available.}
    \label{fig:abalation}
\end{figure*}

\subsection{Role of Structured Knowledge in Model Performance}
Although enhancing prompts with structured knowledge does not consistently improve model performance compared to other prompting techniques, SKP[semantic] leads to slight increases in EMA values (ranging from 0.01\% to 1.32\%) compared to SKP[random], across all models except GPT-2 and Mistral (see Table \ref{tab:main_results}). We identified a subset of MCQs (19.96\%) where all three types of knowledge were available and conducted additional experiments to evaluate the individual contribution of each knowledge type to EMA (we employed SKP[semantic ] prompting). Our results show (see Figure \ref{fig:abalation}, for complete results, see table \ref{tab:additional_results} in Appendix \ref{sec:performance_and_results}) that incorporating each of the three knowledge types separately into prompts leads to very similar EMA values (when averaged across all nine models). Specifically, prompts enhanced only with Wikidata knowledge resulted in an average EMA of 14.57\%, while using only WordNet or only ConceptNet yielded average EMAs of 14.41\% and 14.34\%, respectively.

We also observed that incorporating all three types of knowledge simultaneously into the prompts, compared to using them individually, produced varying results. For example, Falcon, CodeT5 and GPT-3.5-Turbo perform marginally better when a single knowledge type is incorporated into the prompt, compared to including all three knowledge types simultaneously (see Figure \ref{fig:abalation}). Providing FlanT5 with a single knowledge type compared to all three knowledge types contributes to significant increases of percentage points in EMA (WordNet +18.51 , ConceptNet +15.96  and WikiData +7.44). In contrast, GPT-2, Mistral, and Orca perform better when all knowledge types are integrated into the prompt. Notably, Orca demonstrates an average EMA increase of +11.14 percentage points compared to using only a single knowledge source.  

\subsection{Exemplar Quantity vs. Model Performance}
\citet{brown2020language} demonstrated that the accuracy of large language models improves with an increase in the number of exemplars. However, \citet{liu-etal-2022-makes} found that the benefits diminish beyond 20 exemplars in certain cases. Similarly, in our study, increasing exemplars from one to five decreases EMA in six out of nine models (see Table \ref{tab:main_results}), leading us to limit exemplars to a maximum of five.

\begin{table*}[ht!]
\centering
\resizebox{\textwidth}{!}{%
\begin{tabular}{@{}lccccccccc@{}}
\toprule
\multicolumn{1}{c}{} &
  \multicolumn{9}{c}{\textbf{EMA \% of Large Language Model}} \\ \cmidrule(l){2-10} 
\multicolumn{1}{c}{\multirow{-2}{*}{\textbf{Semantic Relationship Type}}} &
  \multicolumn{1}{c}{\textbf{Falcon}} &
  \multicolumn{1}{c}{\textbf{FlanT5}} &
  \multicolumn{1}{c}{\textbf{GPT2}} &
  \multicolumn{1}{c}{\textbf{Mistral}} &
  \multicolumn{1}{c}{\textbf{Orca}} &
  \multicolumn{1}{c}{\textbf{Zephyr}} &
  \multicolumn{1}{c}{\textbf{CodeT5}} &
  \multicolumn{1}{c}{\textbf{CodeParrot}} &
  \textbf{GPT-3.5-Turbo} \\ %
  \midrule
part of &
  \multicolumn{1}{c}{23.50} &
  \multicolumn{1}{c}{\cellcolor[HTML]{FEEAD3}25.05} &
  \multicolumn{1}{c}{20.19} &
  \multicolumn{1}{c}{24.95} &
  \multicolumn{1}{c}{\cellcolor[HTML]{E0FCE0}{\color[HTML]{333333} 23.98}} &
  \multicolumn{1}{c}{9.03} &
  \multicolumn{1}{c}{\cellcolor[HTML]{FEEAD3}{\color[HTML]{333333} 13.88}} &
  \multicolumn{1}{c}{\cellcolor[HTML]{FEEAD3}{\color[HTML]{333333} 0.00}} &
  \cellcolor[HTML]{FEEAD3}{\color[HTML]{333333} 27.18} \\ 
is a &
  \multicolumn{1}{c}{25.78} &
  \multicolumn{1}{c}{29.20} &
  \multicolumn{1}{c}{20.09} &
  \multicolumn{1}{c}{25.50} &
  \multicolumn{1}{c}{22.08} &
  \multicolumn{1}{c}{6.84} &
  \multicolumn{1}{c}{\cellcolor[HTML]{E0FCE0}{\color[HTML]{333333} 19.37}} &
  \multicolumn{1}{c}{\cellcolor[HTML]{FEEAD3}{\color[HTML]{333333} 0.00}} &
  31.34 \\ 
at location &
  \multicolumn{1}{c}{\cellcolor[HTML]{E0FCE0}{\color[HTML]{000000} 29.34}} &
  \multicolumn{1}{c}{31.85} &
  \multicolumn{1}{c}{21.43} &
  \multicolumn{1}{c}{\cellcolor[HTML]{E0FCE0}{\color[HTML]{333333} 27.80}} &
  \multicolumn{1}{c}{\cellcolor[HTML]{FEEAD3}{\color[HTML]{333333} 19.31}} &
  \multicolumn{1}{c}{\cellcolor[HTML]{E0FCE0}{\color[HTML]{333333} 11.20}} &
  \multicolumn{1}{c}{18.73} &
  \multicolumn{1}{c}{\cellcolor[HTML]{E0FCE0}{\color[HTML]{333333} 0.39}} &
  32.24 \\ 
follows &
  \multicolumn{1}{c}{\cellcolor[HTML]{FEEAD3}{\color[HTML]{333333} 22.07}} &
  \multicolumn{1}{c}{28.46} &
  \multicolumn{1}{c}{\cellcolor[HTML]{FEEAD3}{\color[HTML]{333333} 15.43}} &
  \multicolumn{1}{c}{\cellcolor[HTML]{FEEAD3}{\color[HTML]{333333} 22.34}} &
  \multicolumn{1}{c}{23.94} &
  \multicolumn{1}{c}{\cellcolor[HTML]{FEEAD3}{\color[HTML]{333333} 3.19}} &
  \multicolumn{1}{c}{14.36} &
  \multicolumn{1}{c}{0.27} &
  35.11 \\ 
producer &
  \multicolumn{1}{c}{25.94} &
  \multicolumn{1}{c}{\cellcolor[HTML]{E0FCE0}{\color[HTML]{000000} 37.97}} &
  \multicolumn{1}{c}{\cellcolor[HTML]{E0FCE0}{\color[HTML]{333333} 22.19}} &
  \multicolumn{1}{c}{24.87} &
  \multicolumn{1}{c}{23.80} &
  \multicolumn{1}{c}{7.22} &
  \multicolumn{1}{c}{13.90} &
  \multicolumn{1}{c}{\cellcolor[HTML]{FEEAD3}{\color[HTML]{333333} 0.00}} &
  \cellcolor[HTML]{E0FCE0}{\color[HTML]{333333} 53.21} \\ 
\rowcolor[HTML]{EFEFEF} 
\textit{Avg. performance across the above relations} &
  \multicolumn{1}{c}{\cellcolor[HTML]{EFEFEF}\textit{25.33}} &
  \multicolumn{1}{c}{\cellcolor[HTML]{EFEFEF}{\ul \textit{30.51}}} &
  \multicolumn{1}{c}{\cellcolor[HTML]{EFEFEF}\textit{19.87}} &
  \multicolumn{1}{c}{\cellcolor[HTML]{EFEFEF}\textit{25.09}} &
  \multicolumn{1}{c}{\cellcolor[HTML]{EFEFEF}\textit{22.62}} &
  \multicolumn{1}{c}{\cellcolor[HTML]{EFEFEF}\textit{7.50}} &
  \multicolumn{1}{c}{\cellcolor[HTML]{EFEFEF}\textit{16.05}} &
  \multicolumn{1}{c}{\cellcolor[HTML]{EFEFEF}\textit{0.13}} &
  \textit{\textbf{35.82}} \\ \bottomrule
\end{tabular}%
}
\caption{\textbf{Performance of each LLM across semantic relations.} We report the performance of each LLM on the top five semantic relations in the dataset. For each LLM, the relation with the highest performance is highlighted in \colorbox[HTML]{E0FCE0}{green}, while the relation with the lowest performance is highlighted in \colorbox[HTML]{FEEAD3}{orange}. Additionally, the average performance across all five relation types is calculated and highlighted in \colorbox[HTML]{EFEFEF}{grey}.}
\label{tab:relations_results}
\end{table*}

\subsection{Cost of Knowledge Acquisition vs. Model Performance}
In this study, we utilize three types of knowledge to enhance prompts: exemplar knowledge, structured knowledge, and targeted knowledge. Among these, exemplar knowledge has the least acquisition cost since it is readily available from the dataset itself requiring no additional resources. Structured knowledge, on the other hand, is more expensive to acquire because it necessitates accessing external knowledge bases or graphs and filtering knowledge, which incurs computational overhead. Targeted knowledge is the costliest to acquire, as it involves identifying the specific semantic relationship between the question term pairs. This semantic relationship is not always readily available, requiring human annotation (for instance, in our dataset of 15K data points, 1K data points lacked this semantic information, necessitating human annotation). 

As shown in Table \ref{tab:main_results}, targeted knowledge, being the most expensive to acquire, led to the best performance in four models (Falcon, FlanT5, Mistral and GPT-3.5-Turbo) including the peak performance (55\% EMA) from GPT-3.5-Turbo. In contrast, structured knowledge, the second most costly, did not result in any model's best performance. Although exemplar knowledge is the least expensive, three models performed best with it (Zephyr and CodeParrot in Five-shot; CodeT5 in One-shot).

\subsection{Diversity of Semantic Relationships vs. Model Performance}

As elaborated in Section \ref{sec:dataset}, our dataset encompasses 236 unique semantic relation types, with the frequencies of the top five relations detailed in Table \ref{tab:data-stats}. To further elucidate the performance of each LLM, we assess their results on these top five semantic relations using targeted knowledge prompts (refer to Table \ref{tab:relations_results}). Consistent with prior findings, GPT-3.5-Turbo achieves the highest average performance across the top five relations, followed by FlanT5. For both models, the MCQs involving the "part of" relation pose the greatest challenge, whereas the "producer" relation is the easiest to solve. Similarly, across all nine LLMs, the "part of" and "follows" relations emerge as the most difficult, while the "at location" relation proves to be the easiest.

\section{Conclusion and Future Work}
We evaluate nine LLMs on a 15K MCQ dataset to assess their ability to solve proportional analogies using various knowledge-enhanced prompting techniques. Our experiments reveal that LLMs perform best when targeted knowledge is integrated into prompts, outperforming exemplar and structured knowledge.

While several of the models used are instruction-finetuned versions of their base models, they are not specifically finetuned for proportional analogy completion, leaving room for improvement. Additionally, our study focuses on manual prompting techniques, which are brittle; exploring automatic prompting approaches could yield more robust results.


\section{Limitations}
In SKP, knowledge paths may occasionally provide the exact semantic relationship between question term pairs, defined as targeted knowledge. These instances can be classified as SKP with targeted knowledge, but we do not currently verify or adjust for them. Also, in this work, we used manual prompt creation, where slight variations can significantly affect model outputs \cite{zhao2021calibrate}. However, we did not address this variability by testing multiple prompt templates for each prompting technique. Acquiring targeted knowledge is resource-intensive due to the need for manual annotations. Scaling this process is impractical, highlighting the need for automated targeted knowledge acquisition techniques. Not all data points in the dataset include knowledge from ConceptNet, Wikidata, and WordNet due to the incompleteness of these graphs, highlighting the broader challenge of knowledge graph completion.

\section*{Acknowledgments}
We thank anonymous reviewers for their constructive feedback and Anirudh Govil for his valuable input. This work was supported by NSF grant \#2335967: EAGER: Knowledge-guided neurosymbolic AI with guardrails for safe virtual health assistants. Any
opinions, findings, conclusions, or recommendations expressed in this material are those of the
authors and do not necessarily reflect the views of
the funding organization.

\bibliography{custom}

\appendix
\section{Model Details}
\label{sec:model_details}
\textbf{Falcon \cite{almazrouei2023falcon}:} The Falcon model used in this work is the Falcon-7B-Instruct model \footnote{\url{https://huggingface.co/tiiuae/falcon-7b-instruct}} which is a causal decode-only model, instruction finetuned on top of the base Falcon-7B. The fine-tuning dataset is made up of 250M tokens from various conversational datasets (Baize\footnote{\url{https://github.com/project-baize/baize-chatbot/tree/main/data}}), instruction datasets (GPT4All \cite{gpt4all}, GPTeacher\footnote{\url{https://github.com/teknium1/GPTeacher}}) and common crawl data (RefinedWeb \cite{refinedweb})from the web. Falcon-7B tokenizer is used for tokenization. The architecture of Falcon is broadly adapted from GPT3 with changes in positional embeddings used, attention mechanisms used and decoder block architecture.\\

\noindent \textbf{FlanT5 \cite{https://doi.org/10.48550/arxiv.2210.11416}:} We use the FlanT5-XXL version with 11B parameters. This version is based on a pretrained T5 \cite{raffel2020exploring} and instruction finetuned on a mixture of tasks. This model is finetuned specifically with Chain-of-Thought data. \\

\noindent \textbf{GPT2 \cite{radford2019language}: }We use the XL version with 1.5 parameters. The model is pretrained with English language data (40 GB of text from the web) and causal language modeling objective. Interestingly the model is not trained on articles from Wikipedia. \\

\noindent \textbf{Mistral \cite{jiang2023mistral}:} This is a decoder only transformer model and we use the Mistral-7B-Instruct version with 7B parameters. This version is finetuned on publicly available instruction datasets. Mistral introduce Sliding Window Attention, Rolling Buffer Cache and Pre-fill Chunking in its architecture.\\

\noindent \textbf{Orca \cite{orca_mini_7b}:} We employ orca\_mini\_7b, a 7B parameter version of Orca, which is based on OpenLLaMA-7B. The model is trained on datasets with explanation tuning, where the response from the <query, response> pair is augmented with detailed responses from the base (teacher) model \cite{mukherjee2023orca}. The explanation tuning datasets used are WizardLM\footnote{\url{https://github.com/nlpxucan/WizardLM}}, Alpaca dataset \cite{alpaca} and Dolly\footnote{\url{https://github.com/databrickslabs/dolly}} and system prompts are used to elicit step-by-step explanations.
\\

\begin{table*}[!ht]
\centering
\footnotesize
\begin{adjustbox}{width=0.8\textwidth}
\begin{tabular}{@{}lrrrr@{}}
\toprule
\textbf{Model Name}    & \textbf{WD Knowledge Only} & \textbf{CN Knowledge Only} & \textbf{WN Knowledge Only} & \textbf{All Knowledge Available} \\ \midrule
Falcon        & 25.14             & 25.38             & 25.04             & 24.14                   \\
FlanT5        & 19.63             & 28.15             & 30.70              & 12.19                   \\
GPT2          & 4.11              & 2.40               & 1.00                 & 5.84                    \\
Mistral       & 22.84             & 22.07             & 23.34             & 24.11                   \\
Orca          & 14.62             & 9.52              & 9.28              & 22.24                   \\
Zephyr        & 16.16             & 12.49             & 9.75              & 14.76                   \\
CodeT5        & 21.47             & 21.64             & 22.70              & 20.70                    \\
CodeParrot    & 0.00                 & 0.00                 & 0.00                 & 0.00                       \\
GPT-3.5-Turbo & 7.13              & 7.43              & 7.90               & 7.20                     \\ \bottomrule
\end{tabular}
\end{adjustbox}
\caption{\textbf{Performance of models based on provided knowledge types.} Performance values are reported in EMA percentage and calculated using 2995 ($\sim$20\%) data points that had all three knowledge types available.}
\label{tab:additional_results}
\end{table*}

\noindent \textbf{Zephyr\footnote{\url{https://huggingface.co/HuggingFaceH4/zephyr-7b-alpha}}:} We use the Zephyr-7B-alpha with 7B parameters finetuned from Mistral-7B-v0.1. The finetune datasets contain synthetic dialogues ranked by GPT-4 and a prompt completion dataset where completions are ranked by GPT-4.\\

\noindent \textbf{CodeT5 \cite{le2022coderl}:} The CodeT5 model wwe use is codet5-large model with 770M parameters. The model is trained on Masked Span Prediction objective on CodeSearchNet dataset \cite{husain2019codesearchnet}\\

\noindent \textbf{CodeParrot \cite{noauthor_codeparrot_2023}:} We use the 1.5B parameter CodeParrot model based on GPT-2. The model is trained to generate python code on a python files dataset from GitHub\footnote{\url{https://huggingface.co/datasets/codeparrot/codeparrot-clean}}.\\

\noindent \textbf{GPT-3.5-Turbo\footnote{\url{https://platform.openai.com/docs/models/gpt-3-5-turbo}}:} We use OpenAI API to access the model, gpt-3.5-turbo-0125.\\

\begin{figure}[!ht]
    \centering
    \includegraphics[width=\linewidth]{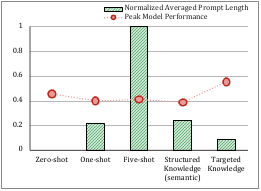}
    \caption{Prompt Lengths vs. Peak Performance.}
    \label{fig:prompt_lengths}
\end{figure}

\section{Implementation Details}
\label{sec:implementation_details}
We use API requests for GPT-3.5-Turbo and checkpoints from Hugging face\footnote{\url{https://huggingface.co/models}} for open-source models. The models are evaluated with following hyper parameter settings, temperature = 0.1, top\_p=0.1 and repetition\_penalty=1.2 to elicit more concrete answers for the MCQs. We use Sentence Transformers\footnote{\url{https://sbert.net/}} to identify semantically similar exemplars and to perform semantic knowledge filtering. We utilize Wikidata knowledge from \cite{wang2021kepler}, ConceptNet knowledge from conceptnet5\footnote{ \url{https://github.com/commonsense/conceptnet5/wiki/Downloads}} and WordNet knowledge from Open English WordNet (2023)\footnote{\url{https://github.com/globalwordnet/english-wordnet?tab=readme-ov-file}}. 

\section{Performance and Additional Results}
\label{sec:performance_and_results}
\subsection{Model Performance vs. Prompt Length (PL)}
We calculated the average prompt lengths across models for each prompting technique (PL for SKP is calculated by averaging SKP[random] and SKP[semantic]) (See Figure \ref{fig:prompt_lengths}). According to \cite{liu-etal-2024-lost}, longer prompts (with important information placed in the middle) tend to negatively affect performance. Based on such literature, one might suggest that Zero-shot prompts yield better results in our study because they are short, but this is not the case. Despite being longer than Zero-shot prompts, a higher peak model performance is achieved by TKP.


\section{Prompts}
\label{sec:prompts}
\begin{figure*}[!ht]
    \centering
    \includegraphics[width=\linewidth]{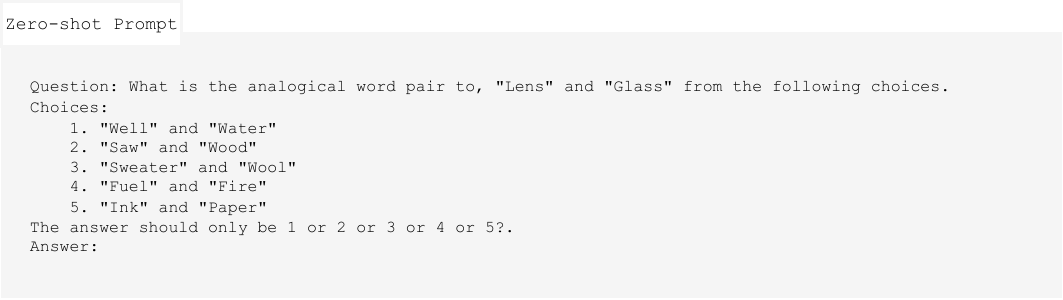}
    \caption{Example of a Zero-shot prompt used on our dataset}
    \label{fig:zero}
\end{figure*}

\begin{figure*}[!ht]
    \centering
    \includegraphics[width=\linewidth]{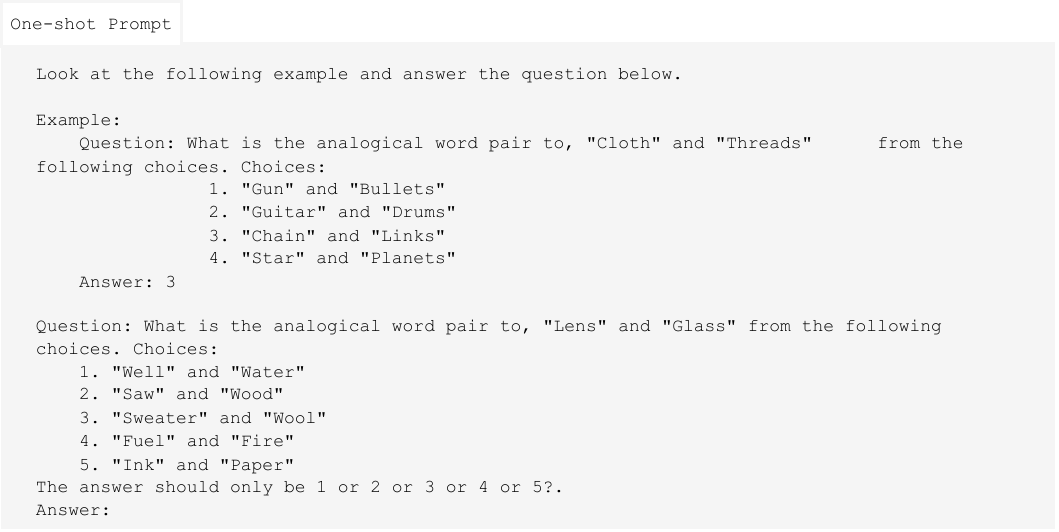}
    \caption{Example of a One-shot prompt used on our dataset}
    \label{fig:one}
\end{figure*}

\begin{figure*}[!ht]
    \centering
    \includegraphics[width=\linewidth]{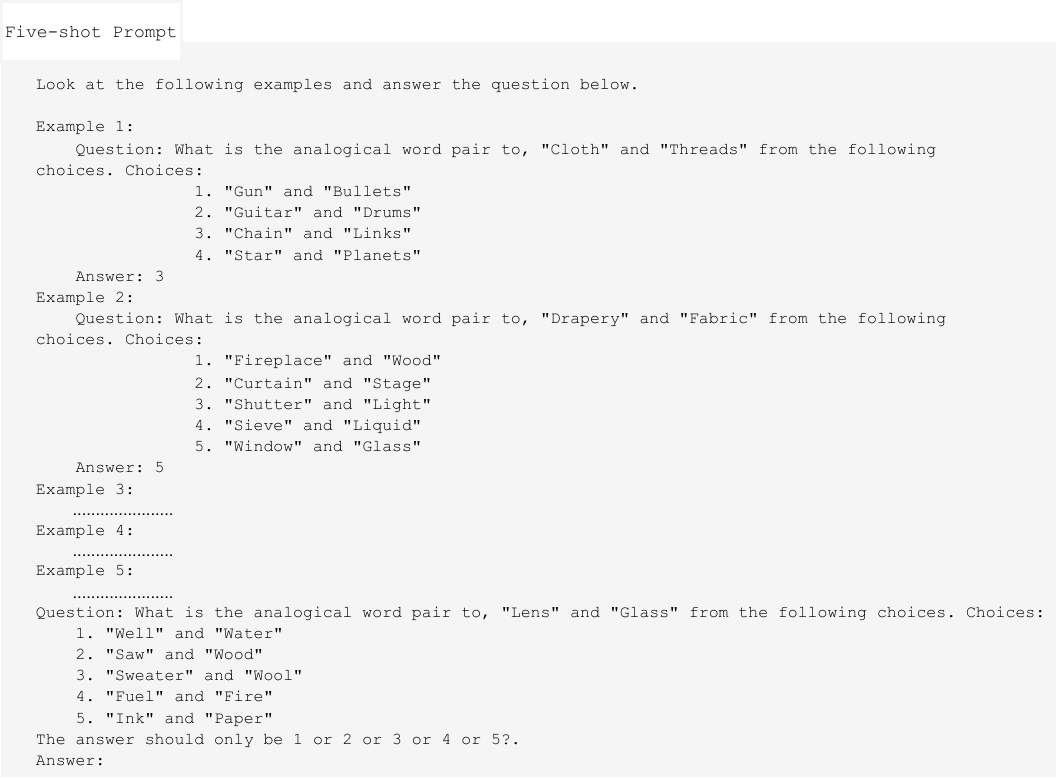}
    \caption{Example of a Five-shot prompt used on our dataset}
    \label{fig:five}
\end{figure*}

\begin{figure*}[!ht]
    \centering
    \includegraphics[width=\linewidth]{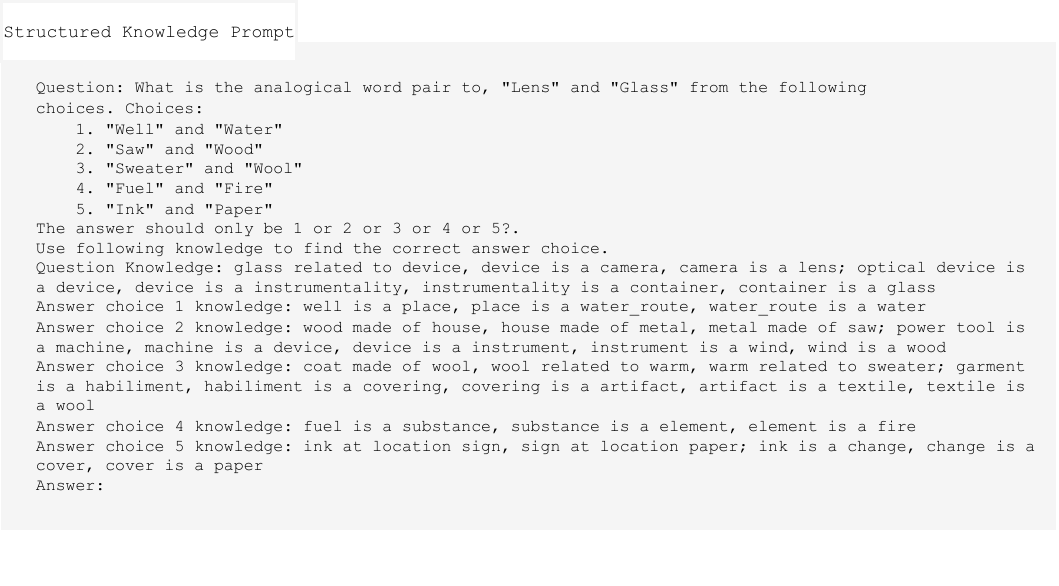}
    \caption{Example of a Structured Knowledge Prompt[semantic] used on our dataset}
    \label{fig:skp}
\end{figure*}

\begin{figure*}[!ht]
    \centering
    \includegraphics[width=\linewidth]{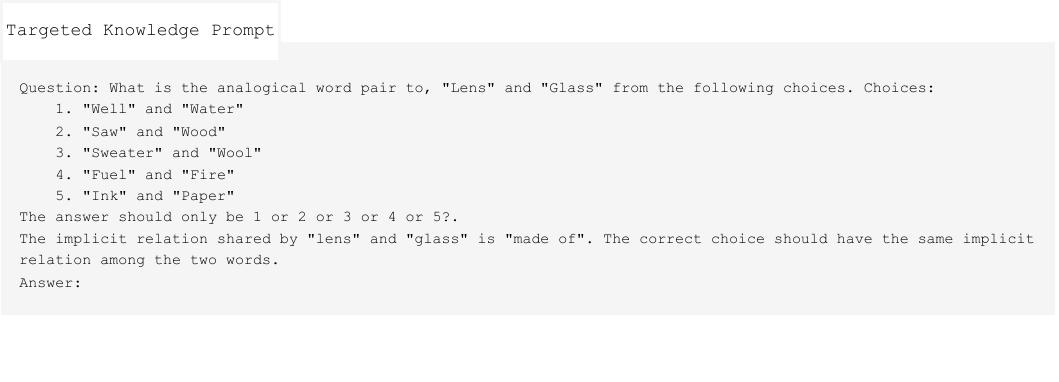}
    \caption{Example of a Targeted Knowledge Prompt used on our dataset}
    \label{fig:tkp}
\end{figure*}

Figues \ref{fig:zero}, \ref{fig:one}, \ref{fig:five}, \ref{fig:skp} and \ref{fig:tkp} illustrates example prompts provided to models. 

\end{document}